%% file: main.tex
\documentclass[10pt,twocolumn,letterpaper]{article}

\usepackage{iccv}
\usepackage{times}
\usepackage{epsfig}
\usepackage{graphicx}
\usepackage{amsmath}
\usepackage{amssymb}

\usepackage{url}            
\usepackage{booktabs}       
\usepackage{amsfonts}       
\usepackage{nicefrac}       
\usepackage{microtype}      
\usepackage{textcomp}
\usepackage{xcolor}
\usepackage{multirow}
\usepackage{flushend}

\usepackage{caption}
\usepackage{subcaption}
\usepackage{paralist}
\usepackage{array}
\usepackage{placeins}
\usepackage{epstopdf}
\usepackage[titletoc,title]{appendix}
\usepackage[T1]{fontenc}
\usepackage[utf8]{inputenc}

\usepackage{pifont}

\usepackage[ruled]{algorithm2e}
\makeatletter
\def\BState{\State\hskip-\ALG@thistlm}
\makeatother

\usepackage{color}
\usepackage{enumitem}
\setitemize{noitemsep,topsep=0pt,parsep=0pt,partopsep=0pt}

\usepackage[pagebackref=true,breaklinks=true,letterpaper=true,colorlinks,bookmarks=false]{hyperref}

\iccvfinalcopy 

\def\iccvPaperID{xxx}

\ificcvfinal\fi
\begin{document}

\title{Boosting Few-Shot Visual Learning with Self-Supervision}

\author{
Spyros Gidaris$^1$, \ \
Andrei Bursuc$^1$, \ \
Nikos Komodakis$^2$, \ \ 
Patrick P\'erez$^1$, \ \
Matthieu Cord$^{1,3}$\\
$^1$valeo.ai \ \ \ \ \ \ \  $^2$LIGM, Ecole des Pont ParisTech\ \ \ \ \ \ \ $^3$Sorbonne Université\\
}

\maketitle

\input{abbrev.tex}
\input{defn.tex}

\begin{abstract}
Few-shot learning and self-supervised learning address different facets of the same problem: how to train a model with little or no labeled data. Few-shot learning aims for optimization methods and models that can learn efficiently to recognize patterns in the low data regime. Self-supervised learning focuses instead on unlabeled data and looks into it for the supervisory signal to feed high capacity deep neural networks. In this work we exploit the complementarity of these two domains and propose an approach for improving few-shot learning through self-supervision. We use self-supervision as an auxiliary task in a few-shot learning pipeline, enabling feature extractors to learn richer and more transferable visual representations while still using few annotated samples. Through self-supervision, our approach can be naturally extended towards using diverse unlabeled data from other datasets in the few-shot setting. We report consistent improvements across an array of architectures, datasets and self-supervision techniques.
\end{abstract}

\input{introduction.tex}

\input{relatedwork.tex}

\input{methodology.tex}

\input{experiments.tex}

\input{conclusions.tex}

\clearpage
{\small
\bibliographystyle{ieee}
\bibliography{main}
}

\appendix
\input{appendix.tex}
\end{document}

%% file: abbrev.tex
\newcommand{\tran}{^\top}
\newcommand{\mtran}{^{-\top}}
\newcommand{\zcol}{\mathbf{0}}
\newcommand{\zrow}{\zcol\tran}

\newcommand{\ind}{\mathbbm{1}}
\newcommand{\expect}{\mathbb{E}}
\newcommand{\nat}{\mathbb{N}}
\newcommand{\zahl}{\mathbb{Z}}
\newcommand{\real}{\mathbb{R}}
\newcommand{\proj}{\mathbb{P}}
\newcommand{\prob}{\mathbf{Pr}}
\newcommand{\softmax}{\operatorname{softmax}}

\newcommand{\defn}{\mathrel{:=}}

\newcommand{\vp}{\mathbf{p}}
\newcommand{\vq}{\mathbf{q}}
\newcommand{\vu}{\mathbf{u}}
\newcommand{\vv}{\mathbf{v}}
\newcommand{\vw}{\mathbf{w}}
\newcommand{\vx}{\mathbf{x}}
\newcommand{\vy}{\mathbf{y}}
\newcommand{\vz}{\mathbf{z}}

\newcommand{\mv}[1]{\ensuremath{\bm{#1}}} 
\newcommand{\mm}[1]{\ensuremath{\bm{#1}}} 

\newcommand{\Set}[2]{\{\, #1 \mid #2 \, \}}


\newcommand{\few}{\mathrm{few}}
\newcommand{\self}{\mathrm{self}}

\newcommand{\mini}{{MiniImagenet}}
\newcommand{\tmini}{{tiered-MiniImagenet}}
\newcommand{\fscifar}{{CIFAR-FS}}

\makeatletter
\DeclareRobustCommand\onedot{\futurelet\@let@token\@onedot}
\def\@onedot{\ifx\@let@token.\else.\null\fi\xspace}
\def\eg{\emph{e.g}\onedot} \def\Eg{\emph{E.g}\onedot}
\def\ie{\emph{i.e}\onedot} \def\Ie{\emph{I.e}\onedot}
\def\cf{\emph{cf}\onedot} \def\Cf{\emph{Cf}\onedot}
\def\etc{\emph{etc}\onedot} \def\vs{\emph{vs}\onedot}
\def\wrt{w.r.t\onedot} \def\dof{d.o.f\onedot}
\def\etal{\emph{et al}\onedot}
\makeatother
\newcommand{\secref}[1]{\S\ref{#1}}

%% file: defn.tex
\makeatletter
\newcommand*{\inlineequation}[2][]{%
  \begingroup
    \refstepcounter{equation}%
    \ifx\\#1\\%
    \else
      \label{#1}%
    \fi
    \relpenalty=10000 %
    \binoppenalty=10000 %
    \ensuremath{%
      #2%
    }%
    ~\@eqnnum
  \endgroup
}
\makeatother

\newcommand{\Strn}{\ensuremath{S_a}}
\newcommand{\Stst}{\ensuremath{S_b}}

%% file: introduction.tex
\section{Introduction}

Deep learning based methods have achieved impressive results on various image understanding tasks, such as image classification~\cite{he2016deep, krizhevsky2012imagenet, simonyan2014very, szegedy2015going}, object detection~\cite{ren2015faster}, or semantic segmentation~\cite{chen2018deeplab}.
However, in order to successfully learn these tasks, 
such models need to access large volumes of manually labeled training data.
If not, the trained models might suffer from poor generalization performance on the test data. 
In image classification for instance, learning to recognize reliably a set of categories with convolutional neural networks (convnets) requires hundreds or thousands of training examples per class.
In contrast, humans are perfectly capable of learning new visual concepts from only one or few examples, generalizing without difficulty to new data.
The aim of \emph{few-shot learning}~\cite{fe2003bayesian, fei2006one, koch2015siamese, lake2011one, vinyals2016matching} is to endow artificial perception systems with a similar ability, especially with the help of modern deep learning. 

Hence, the goal of few-shot visual learning is to devise recognition models
that are capable of efficiently learning to recognize a set of classes despite the fact that there are available very few training examples for them (\eg, only 1 or 5 examples per class).
In order to avoid overfitting due to data scarcity,
few-shot learning algorithms rely on transfer learning techniques and have two learning stages.
During a first stage, the model is usually trained using a different set of classes, called \textit{base classes}, 
which is associated to a large set of annotated training examples.
The goal of this stage is to let the few-shot model acquire transferable visual analysis abilities, typically in the form of learned representations, that are mobilized in the second stage.
In this subsequent step, 
the model indeed learns to recognize \textit{novel classes}, unseen during the first learning stage, using only a few training examples per class.

\begin{figure*}[!ht]
\renewcommand{\captionfont}{\small}
\centering
\includegraphics[width=0.85\linewidth]{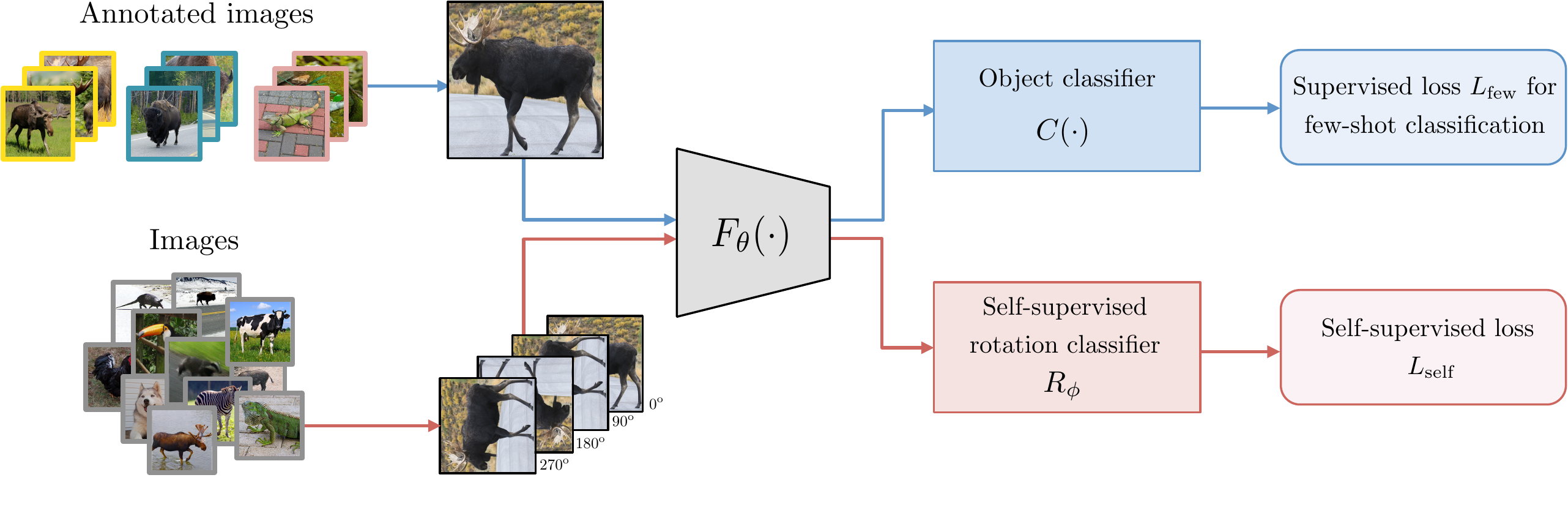}
\vspace{-10pt}
\caption{\textbf{Combining self-supervised image rotation prediction and supervised base class recognition in first learning stage of a few-shot system.} We train the feature extractor $F_{\theta}(\cdot)$ with both annotated (top branch) and non-annotated (bottom branch) data in a multi-task setting. We use the annotated data to train the object classifier $C(\cdot)$ with the few-shot classification loss $L_{\mathrm{few}}$. For the self-supervised task, we sample images from the annotated set (and optionally from a different set of non-annotated images). Here, we generate four rotations for each input image, process them with $F_{\theta}(\cdot)$ and train the rotation classifier $R_{\phi}$ with the self-supervised loss $L_{\mathrm{self}}$. The pipeline for relative patch location self-supervision is analogue to this one.}
\label{fig:self_rotation}
\vspace{-10pt}
\end{figure*}

Few-shot learning relates with self-supervised representation learning~\cite{doersch2015unsupervised, dosovitskiy2014discriminative, gidaris2018unsupervised, larsson2016learning, noroozi2016unsupervised, zhang2016colorful}.
The latter is a form of unsupervised learning that trains a model on an annotation-free pretext task defined using only the visual information present in images.
The purpose of this self-supervised task is to make the model learn image representations that would be useful when transferred to other image understanding tasks.
For instance, in the seminal work of Doersch \etal~\cite{doersch2015unsupervised},
a network, by being trained on the self-supervised task of predicting the relative location of image patches, 
manages to learn image representations that are successfully transferred to the vision tasks of object recognition, object detection, and semantic segmentation.
Therefore, as in few-shot learning, self-supervised methods also have two learning stages,
the first that learns image representations with a pretext self-supervised task, and the second that adapts those representations to the actual task of interest.
Also, both learning approaches try to limit the dependence of deep learning methods on the availability of large amounts of labeled data.

Inspired by the connection between few-shot learning and self-supervised learning,
we propose to combine the two methods to improve the transfer learning abilities of few-shot models.
More specifically, we propose to add a self-supervised loss to the training loss that a few-shot model minimizes during its first learning stage (see  Figure~\ref{fig:self_rotation}).
Hence, we artificially augment the training task(s) that a few-shot model solves during its first learning stage.
We argue that this task augmentation 
forces the model to learn a more diversified set of image features, 
and this in turn improves its ability to adapt to novel classes with few training data.
Moreover, since self-supervision does not require data annotations, one can include extra unlabeled data to the first learning stage. By extending the size and variety of training data in this manner, one might expect to learn richer image features and to get further performance gain in few-shot learning. At the extreme, using only unlabeled data in the first learning stage, thus removing the use of base classes altogether, is also appealing. We will show that both these semi-supervised and unsupervised regimes can be indeed put at work for few-shot recognition thanks to self-supervised tasks.

In summary, the contributions of our work are: 
\textbf{(1)}
We propose to weave self-supervision into the training objective of few-shot learning algorithms.
The goal is to boost the ability of the latter to adapt to novel classes with few training data.
\textbf{(2)}
We  study the impact of the added self-supervised loss by performing exhaustive quantitative experiments on \mini, \fscifar, and \tmini~few-shot datasets.
In all of them self-supervision improves the few-shot learning performance leading to state-of-the-art results.
\textbf{(3)}
Finally, we extend the proposed few-shot recognition framework to semi-supervised and unsupervised setups, getting further performance gain in the former, and
showing with the latter that our framework can be used for evaluating and comparing self-supervised representation learning approaches on few-shot object recognition.

%% file: relatedwork.tex
\section{Related work} \label{sec:related_work}

\paragraph{Few-shot learning.} 
There is a broad array of few-shot learning approaches, including, among many: gradient descent-based approaches~\cite{andrychowicz2016learning,finn2017model, nichol2018reptile, ravi2016optimization}, which learn how to rapidly adapt a model to a given few-shot recognition task via a small number of gradient descent iterations; 
metric learning based approaches that learn a distance metric between a query, \ie, test image, and a set of support images, \ie, training images, of a few-shot task~\cite{koch2015siamese, snell2017prototypical, vinyals2016matching, wang2018low, yang2018learning}; methods learning to map a test example to a class label by accessing memory modules that store training examples for that task~\cite{garcia2017few, kaiser2017learning, mishra2018simple,  munkhdalai2017meta, santoro2016meta}; approaches that learn how to generate the weights of a classifier~\cite{gidaris2018dynamic, gomez2005evolving, qi2018low, qiao2017few} or of a multi-layer neural network~\cite{bertinetto2016, ha2016hypernetworks, han2018, wang2017} for the new classes given the few available training data for each of them; methods that ``hallucinate'' additional examples of a class from a reduced amount of data~\cite{hariharan2017low, wang2018low}.

In our work we consider two approaches from the metric learning category, namely \emph{Prototypical Networks}~\cite{snell2017prototypical} and \emph{Cosine Classifiers}~\cite{gidaris2018dynamic, qi2018low} for their simplicity and flexibility. Nevertheless, the proposed auxiliary self-supervision is compatible with several other few-shot classification solutions.

\paragraph{Self-supervised learning.} This is a recent paradigm which is mid-way between unsupervised and supervised learning, and aims to mitigate the challenging need for large amounts of annotated data. 
Self-supervised learning defines an annotation-free pretext task, 
in order to provide a surrogate supervision signal for feature learning. 
Predicting the colors of images~\cite{larsson2016learning, zhang2016colorful}, 
the relative position of image patches~\cite{doersch2015unsupervised, noroozi2016unsupervised}, 
the random rotation that has been applied to an image \cite{gidaris2018unsupervised}, 
or the missing central part of an image~\cite{pathak2016context}, are some of the many methods~\cite{godard2017unsupervised, lee2017unsupervised, misra2016shuffle, vondrick2018tracking, zhou2017unsupervised} for self-supervised feature learning.
The intuition is that, by solving such tasks, the trained model extracts semantic features that can be useful for other downstream tasks. 
In our case, we consider a multi-task setting where we train the backbone convnet using joint supervision from the supervised end-task and an auxiliary self-supervised pretext task. 
Unlike most multi-task settings aiming at good results on all tasks simultaneously~\cite{kokkinos2017ubernet}, we are interested in improving performance on the main task only by leveraging supervision from the surrogate task, as also shown in~\cite{mordan2018revisiting}.  
We expect that, in a few-shot setting where squeezing out generalizable features from the available data is highly important, the use of self-supervision as an auxiliary task will bring improvements.
Also, related to our work,
Chen \etal~\cite{chen2018self} recently added rotation prediction self-supervision to generative adversarial networks~\cite{goodfellow2014generative} leading to significant quality improvements of the synthesized images.

%% file: methodology.tex
\section{Methodology} \label{sec:methodology}

As already explained, few-shot learning algorithms have two learning stages and two corresponding sets of classes.
Here, we define as $D_{b} = \{(\vx,y)\} \subset I\times Y_b$ 
the training set of \textit{base classes}
used during the first learning stage, where $\vx\in I$ is an image with label $y$ in label set $Y_b$ of size $N_b$. 
Also, we define as 
$D_{n} = \{(\vx, y)\} \subset I \times Y_n$ 
the training set of $N_n$ \textit{novel classes} 
used during the second learning stage, where each class has $K$ samples ($K=1$ or $5$ in benchmarks). One talks about $N_n$-way $K$-shot learning. 
Importantly, the label sets $Y_{n}$ and $Y_{b}$ are disjoint.

In the remainder of this section, 
we first describe in~\S\ref{sec:fewshot} the two standard few-shot learning methods that we consider and 
introduce in~\S\ref{sec:selfsupervision} the proposed method to boost their performance with self-supervision.

\subsection{Explored few-shot learning methods} \label{sec:fewshot}

The main component of all few-shot algorithms is a feature extractor $F_{\theta}(\cdot)$, which is a convnet with parameters $\theta$.
Given an image $\vx$, the feature extractor will output a $d$-dimensional feature $F_{\theta}(\vx)$.
In this work we consider two few-shot algorithms, Prototypical Networks (PN)~\cite{snell2017prototypical} and Cosine Classifiers (CC)~\cite{gidaris2018dynamic,qi2018low}, described below. They are fairly similar, with their main difference lying in the first learning stage: only CC learns actual base classifiers along with the feature extractor, while PN simply relies on class-level averages. 

\paragraph{Prototypical Networks (PN)~\cite{snell2017prototypical}.} 

During the first stage of this approach, the feature extractor $F_{\theta}(\cdot)$ 
is learned on sampled few-shot classification sub-problems that are analogue to the targeted few-shot classification problem.
In each training episode of this learning stage, a subset $Y_*\subset Y_b$ of $N_*$ base classes are sampled (they are called ``support classes'') and, for each of them, $K$ training examples are randomly picked from within $D_b$. This yields a training set $D_*$. Given current feature extractor $F_{\theta}$, the average feature for each class $j\in Y_*$, its ``prototype'', is computed as
\begin{equation}
    \vp_j = \frac{1}{K} \sum_{\vx \in X_*^j} F_{\theta}(\vx),~\mathrm{with}~X_*^j = \Set{\vx}{(\vx,y)\in D_*,~y = j}
    \label{eq:proto}
\end{equation}
and used to build a simple similarity-based classifier.   
Then, given a new image $\vx_q$ from a support class but different from samples in $D_{*}$, the classifier outputs for each class $j$ the normalized classification score
\begin{equation}
C^j(F_{\theta}(\vx_q); D_*) = \mathop{\softmax_{j}}
\Big[\mathrm{sim}\big(F_{\theta}(\vx_q), \vp_i \big)_{i\in Y_*} \Big],
\end{equation}
where $\mathrm{sim}(\cdot,\cdot)$ is a similarity function, which may be cosine similarity or negative squared Euclidean distance.  So, in practice, the image $\vx_q$ will be classified to its closest prototype. 
Note that the classifier is conditioned on $D_*$ in order to compute the class prototypes. 
The first learning stage finally amounts to iteratively minimizing the following loss \wrt $\theta$:
\begin{equation} \label{eq:pn_classifier_loss}
L_{\few}(\theta; D_b) = 
\mathop{\expect}_{\tiny\begin{matrix}D_*\sim D_b\\(\vx_q, y_q)\end{matrix}} 
\Big[-\log C^{y_q}(F_{\theta}(\vx_q); D_*)\Big] \textrm{,}
\end{equation}
where $(\vx_q, y_q)$ is a training sample from a support class defined in $D_*$ but different from images in $D_*$.

In the second learning stage, the feature extractor $F_{\theta}$ is frozen and the classifier of novel classes is simply defined as $C(\cdot;D_n)$, with prototypes defined as in (\ref{eq:proto}) with $D_* =D_n$.

\paragraph{Cosine Classifiers (CC) ~\cite{gidaris2018dynamic, qi2018low}.}

In CC few-shot learning, the first stage trains the feature extractor $F_{\theta}$ together with a cosine-similarity based classifier on the (standard) supervised task of classifying the base classes. Denoting $W_b = [\vw_1, ..., \vw_{N_b}]$ the matrix of the $d$-dimensional classification weight vectors, the normalized score for an input image $\vx$ reads
\begin{equation}
    C^{j}(F_{\theta}(\vx);W_b) = 
    \mathop{\softmax_{j}}\Big[\gamma\cos\big(F_{\theta}(\vx),\vw_i\big)_{i\in Y_b}\Big],
\label{eq:cos_classifier}
\end{equation}
where $\cos(\cdot,\cdot)$ is the cosine operation between two vectors, and the scalar $\gamma$ is the inverse temperature parameter of the softmax operator.\footnote{Specifically, $\gamma$ controls the peakiness of the probability distribution generated by the softmax operator~\cite{hinton2015distilling}.} 

The first learning stage aims at minimizing \wrt $\theta$ and $W_b$ the negative log-likelihood loss:
\begin{equation} \label{eq:cos_classifier_loss}
L_{\few}(\theta, W_b; D_b) = \mathop{\expect}_{(\vx, y) \sim D_b} \big[-\log C^{y}(F_{\theta}(\vx);W_b)\big] \textrm{.}
\end{equation}

One of the reasons for using the cosine-similarity based classifier instead of the standard dot-product based one, 
is that the former learns feature extractors that reduce intra-class variations and thus can generalize better on novel classes. By analogy with PN, the weight vectors $\vw_j$'s can be interpreted as learned prototypes for the base classes, to which input image features are compared for classification.  

As with PN, the second stage boils down to computing one representative feature $\vw_j$ for each new class by simple averaging of associated $K$ samples in $D_n$, and to define the final classifier $C(.;[\vw_1\cdots \vw_{N_n}])$ the same way as in (\ref{eq:cos_classifier}).

\subsection{Boosting few-shot learning via self-supervision} \label{sec:selfsupervision}

A major challenge in few-shot learning is encountered during the first stage of learning. How to make the feature extractor learn image features that can be readily exploited for novel classes with few training data during the second stage? With this goal in mind, we propose to leverage the recent progress in  self-supervised feature learning to further improve current few-shot learning approaches. 

Through solving a non-trivial pretext task that can be trivially supervised, such as recovering the colors of images from their intensities, a network is encouraged to learn rich and generic image features that are transferable to other downstream tasks such as image classification. 
In the first stage of few-shot learning, we propose to extend the training of the  feature extractor $F_{\theta}(.)$ by including such a self-supervised task besides the main task of recognizing base classes.

We consider two ways for incorporating self-supervision into few-shot learning algorithms:
(1) by using an auxiliary loss function based on a self-supervised task, and 
(2) by exploiting unlabeled data in a semi-supervised way during training.
In the following we will describe the two techniques.

\subsubsection{Auxiliary loss based on self-supervision}
\label{sec:aux_losses}

We incorporate self-supervision to a few-shot learning algorithm
by adding an auxiliary self-supervised loss during its first learning stage.
More formally, 
let $L_{\self}(\theta, \phi; X_b)$ be the self-supervised loss applied to the set $X_b = \Set{\vx}{ (\vx,y) \in D_b}$ of training examples in $D_b$ deprived of their class labels. 
The loss $L_{\self}(\theta, \phi; X_b)$ is a function of the parameters $\theta$
of the feature extractor and of the parameters $\phi$ of a network only dedicated to the self-supervised task. 
The first training stage of few-shot learning now reads 
\begin{equation} \label{eq:total_loss}
 \min_{\theta, [W_b], \phi} L_{\few}(\theta, [W_b]; D_b) + \alpha  L_{\self}(\theta, \phi; X_b) \textrm{ ,}
\end{equation}
where $L_{\few}$ stands either for the PN few-shot loss (\ref{eq:pn_classifier_loss}) or for the CC one (\ref{eq:cos_classifier_loss}), with additional argument $W_b$ in the latter case (hence bracket notation). The positive hyper-parameter $\alpha$ controls the importance of the self-supervised term\footnote{In our experiments, we use $\alpha=1.0$.}.
An illustration of the approach is provided in Figure~\ref{fig:self_rotation}.

For the self-supervised loss, we consider two tasks in the present work: predicting the rotation incurred by an image, ~\cite{gidaris2018unsupervised}, which is simple and readily incorporated into a few-shot learning algorithm;
predicting the relative location of two patches from the same image~\cite{doersch2015unsupervised}, a seminal task in self-supervised learning. In a recent study, both methods have been shown to achieve state-of-the-art results~\cite{kolesnikov2019revisiting}. 

\paragraph{Image rotations.} 

In this task, the convnet must recognize among four possible 2D rotations in $\mathcal{R} = \{ 0^{\circ}, 90^{\circ}, 180^{\circ}, 270^{\circ} \}$ the one applied to an image (see Figure~\ref{fig:self_rotation}).
Specifically, given an image $\vx$, we first create its four rotated copies $\Set{\vx^r}{ r \in \mathcal{R}}$, 
where $\vx^r$ is the image $\vx$ rotated by $r$ degrees.
Based on the features $F_{\theta}(\vx^{r})$ extracted from such a rotated image, the new network $R_{\phi}$ attempts to predict the rotation class $r$. 
Accordingly, the self-supervised loss of this task is defined as:
\begin{equation} \label{eq:rotation_loss}
L_{\self}(\theta, \phi; X) = \mathop{\mathbb{E}}_{\vx \sim X} \Big[\sum_{\forall r \in \mathcal{R}} -\log R_{\phi}^r\big( F_{\theta}(\vx^r) \big)\Big] \textrm{ ,}
\end{equation}
where $X$ is the original training set of non-rotated images 
and $R_{\phi}^r(\cdot)$ is the predicted normalized score for rotation $r$.
Intuitively, in order to do well for this task the model should reduce the bias towards \emph{up-right} oriented images, typical for ImageNet-like datasets, and learn more diverse features to disentangle classes in the low-data regime.

\paragraph{Relative patch location.} 

Here, we create random pairs of patches from an image and then predict, among eight possible positions, the location of the second patch \wrt to the first, \eg, ``on the left and above'' or ``on the right and below''.
Specifically, given an image $\vx$, 
we first divide it into $9$ regions over a $3 \times 3$ grid and sample a patch within each region. Let's denote $\bar\vx^0$ the central image patch, and $\bar\vx^1\cdots\bar\vx^8$ its eight neighbors lexicographically ordered.
We compute the representation of each patch\footnote{If the architecture of $F_{\theta}(\cdot)$ is fully convolutional, we can apply it to both big images and smaller patches.} and then generate patch feature pairs $\big(F_{\theta}(\bar\vx^0), F_{\theta}(\bar\vx^p)\big)$ by concatenation. We train a fully-connected network $P_{\phi}(\cdot,\cdot)$ to predict the position of $\bar\vx^p$ from each pair.

The self-supervised loss of this task is defined as:
\begin{equation} \label{eq:position_loss}
L_{\self}(\theta, \phi; X) = \mathop{\expect}_{\vx \sim X} \Big[\sum_{p=1}^8 -\log P_{\phi}^p\big( F_{\theta}(\bar\vx^0), F_{\theta}(\bar\vx^p) \big)\Big] \textrm{ ,}
\end{equation}
where $X$ is a set of images and $P_{\phi}^p$ is the predicted normalized score for the relative location $p$.
Intuitively, a good model on this task should somewhat recognize objects and parts, even in presence of occlusions and background clutter.
Note that, in order to prevent the model from learning low-level image statistics such as chromatic aberration~\cite{doersch2015unsupervised}, the patches are preprocessed with aggressive color augmentation (\ie, converting to grayscale with probability $0.66$ and normalizing the pixels of each patch individually to have zero mean and unit standard deviation).

\subsubsection{Semi-supervised few-shot learning} 

The self-supervised term $L_{\self}$ in the training loss (\ref{eq:total_loss}) does not depend on class labels. 
We can easily extend it to learn as well from additional unlabeled data. 
Indeed, if a set $X_u$ of unlabeled images is available besides $D_b$, 
we can make the self-supervised task benefit from them by redefining the first learning stage as:
\begin{equation} \label{eq:total_loss_semisup}
\min_{\theta, [W_b], \phi} L_{\few}(\theta, [W_b]; D_b) + \alpha \cdot L_{\self}(\theta, \phi; X_b \cup X_u) \textrm{ .}
\end{equation}
By training the feature extractor $F_{\theta}$ to also minimize the self-supervised loss on these extra unlabeled images, we open up its visual scope with the hope that this will further improve its ability to accommodate novel classes with scarce data.
This can be seen as a semi-supervised training approach for few-shot algorithms.
An interesting aspect of this semi-supervised training approach is that it does not require the extra unlabeled data to be from the same (base) classes as those in labeled dataset $D_b$. Thus, it is much more flexible \wrt the source of the unlabeled data than standard semi-supervised approaches.

%% file: experiments.tex
\section{Experimental Results} \label{sec:experiments}

In this section we evaluate self-supervision as auxiliary loss function in~\S\ref{sec:aux_experiments} and in~\S\ref{sec:semi_experiments} as a way of exploiting unlabeled data in semi-supervised training.
Finally, in~\S\ref{sec:self_experiments} we use the few-shot object recognition task for evaluating self-supervised methods.

\paragraph{Datasets.}
We perform experiments on three few-shot datasets, 
MiniImageNet~\cite{vinyals2016matching},
\emph{tiered}-MiniImageNet~\cite{ren2018meta} and
CIFAR-FS~\cite{bertinetto2018meta}.
\textbf{MiniImageNet} consists of $100$ classes randomly picked from the ImageNet dataset~\cite{russakovsky2015imagenet} (\ie, $64$ base classes, $16$ validation classes, and $20$ novel test classes);
each class has $600$ images with size $84 \times 84$ pixels. 
\textbf{\emph{tiered}-MiniImageNet} consists of $608$ classes randomly picked from ImageNet~\cite{russakovsky2015imagenet} (\ie, $351$ base classes, $97$ validation classes, and $160$ novel test classes); 
in total there are $779,165$ images again with size $84 \times 84$. 
Finally, \textbf{CIFAR-FS} is a few-shot dataset created by dividing the $100$ classes of CIFAR-100 into
$64$ base classes, $16$ validation classes, and $20$ novel test classes. 
The images in this dataset have size $32 \times 32$ pixels.

\paragraph{Evaluation metrics.}
Few-shot classification algorithms are evaluated based on the classification performance in their second learning stage (when the learned classifier is applied to test images from the novel classes).
More specifically, a large number of $N_n$-way $K$-shot tasks are sampled from the available set of novel classes. 
Each task is created by randomly selecting $N_n$ novel classes from the available test (validation) classes and then within the selected images randomly selecting $K$ training images and $M$ test images per class (making sure that train and test images do not overlap).
The classification performance is measured on the $N_n \times M$ test images and is averaged over all the sampled few-shot tasks. 
Except otherwise stated, for all experiments we used $M = 15$, $N_n =5$, and $K=1$ or $K=5$ (1-shot and 5-shot settings respectively). 

\subsection{Implementation details} \label{sec:impl_details}

\paragraph{Network architectures.}
We conduct experiments with $3$ different feature extractor  architectures $F_{\theta}$, Conv-4-64, Conv-4-512, and WRN-28-10. 
\textbf{Conv-4-64}~\cite{vinyals2016matching} consists of $4$ convolutional blocks each implemented with a $3 \times 3$ convolutional layer followed by BatchNorm~\cite{ioffe2015batch} + ReLU + $2 \times 2$ max-pooling units~\cite{vinyals2016matching}. All blocks of Conv-4-64 have $64$ feature channels.
The final feature map has size $5 \times 5 \times 64$ and is flattened into a final $1600$-dimensional feature vector.
\textbf{Conv-4-512} is derived from Conv-4-64 by gradually increasing its width across layers leading to $96$, $128$, $256$, and $512$ feature channels for its 4 convolutional blocks respectively. Therefore, its output feature vector has $5 \times 5 \times 512 = 12,800$ dimensions after flattening.
Finally, \textbf{WRN-28-10} is a $28$-layer Wide Residual Network~\cite{zagoruyko2016wide} with width factor $10$. 
Its output feature map has size $10 \times 10 \times 640$ which after global average pooling creates a $640$-dimensional feature vector.

The network $R_{\phi}(\cdot)$ specific to the rotation prediction task
gets as input the output feature maps of $F_{\theta}$ and is implemented as a convnet. 
Given two patches, the network $P_{\phi}(\cdot, \cdot)$ specific to the relative patch location task
gets the concatenation of their feature vectors extracted with $F_{\theta}$ as input, 
and forwards it to two fully connected layers.

For further architecture details see Appendix~\ref{sec:appendix_network_architectures}.

\paragraph{Training optimization routine for fist learning stage.}
The training loss is optimized with mini-batch stochastic gradient descent (SGD). For the labeled data we apply both recognition $L_{\few}$ and self-supervised $L_{\self}$ losses. 
For the semi-supervised training, 
at each step we sample mini-batches that consist of labeled data, for which we use both losses, and unlabeled data, in which case we apply only $L_{\self}$.
Learning rates, number of epochs, and batches sizes, were cross-validated on the validation sets.
For further implementation details see Appendices~\ref{sec:appendix_selfsupervision_training} and~\ref{sec:appendix_1step_training}.

\paragraph{Implementation of relative patch location task.}
Due to the aggressive color augmentation of the patches in the patch localization task, and the fact that the patches are around 9 times smaller than the original images, 
the data distribution that the feature extractor ``sees'' from them is very different from that of the images. 
To overcome this problem we apply an extra auxiliary classification loss to the features extracted from the patches.
Specifically, during the first learning stage of CC we merge the features $F_{\theta}(\bar{\vx}^p)$ of the 9 patches of an image (\eg, with concatenation or averaging) and then apply the cosine classifier~(\ref{eq:cos_classifier}) to the resulting feature (this classifier does not share its weight vectors with the classifier applied to the original images features). 
Note that this patch based auxiliary classification loss has the same weight as the original classification loss $L_{\few}$.
Also, during the second learning stage we do not use the patch based classifier.


\subsection{Self-supervision as auxiliary loss function} \label{sec:aux_experiments}
\begin{table}[t]
\centering
\renewcommand{\figurename}{Table}
\renewcommand{\captionlabelfont}{\bf}
\renewcommand{\captionfont}{\small}
{\setlength{\extrarowheight}{2pt}\footnotesize
{
\begin{tabular}{ >{\hspace{-0.2em}} l <{\hspace{-0.2em}} | >{\hspace{-0.2em}} c <{\hspace{-0.2em}} |  >{\hspace{-0.2em}} l <{\hspace{-0.2em}} | >{\hspace{-0.2em}} l <{\hspace{-0.2em}}}
\toprule
\multicolumn{1}{l|}{Model}  & Backbone & \multicolumn{1}{c|}{1-shot} & \multicolumn{1}{c}{5-shot}\\
\midrule
\; CC       & \multirow{2}{*}{Conv-4-64}   & 54.31 $\pm$ 0.42\% & 70.89 $\pm$ 0.34\%\\
\; CC+rot   &                              &\textbf{54.83} $\pm$ 0.43\% & \textbf{71.86} $\pm$ 0.33\%\\
\midrule
\; CC       & \multirow{2}{*}{Conv-4-512}  & 55.68 $\pm$ 0.43\% & 73.19 $\pm$ 0.33\%\\
\; CC+rot    &                              & \textbf{56.27} $\pm$ 0.43\% & \textbf{74.30} $\pm$ 0.33\%\\
\midrule
\; CC       & \multirow{2}{*}{WRN-28-10} &  61.09 $\pm$ 0.44\% & 78.43 $\pm$ 0.33\%\\
\; CC+rot   &                            &  \textbf{62.93} $\pm$ 0.45\% & \textbf{79.87} $\pm$ 0.33\%\\
\midrule
\; PN       & \multirow{2}{*}{Conv-4-64} &  52.20 $\pm$ 0.46\% & 69.98 $\pm$ 0.36\%\\
\; PN+rot    &                            &  \textbf{53.63} $\pm$ 0.43\% & \textbf{71.70} $\pm$ 0.36\%\\
\midrule
\; PN       & \multirow{2}{*}{Conv-4-512} & 54.60 $\pm$ 0.46\% & 71.59 $\pm$ 0.36\%\\
\; PN+rot   &                             & \textbf{56.02} $\pm$ 0.46\% & \textbf{74.00} $\pm$ 0.35\%\\
\midrule
\; PN       & \multirow{2}{*}{WRN-28-10} & 55.85 $\pm$ 0.48\% & 68.72 $\pm$ 0.36\%\\
\; PN+rot   &                            & \textbf{58.28} $\pm$ 0.49\% & \textbf{72.13} $\pm$ 0.38\%\\
\bottomrule
\end{tabular}}}
\caption{\textbf{Rotation prediction as auxiliary loss on MiniImageNet.} Average 5-way classification accuracies for the novel classes on the test set of MiniImageNet with $95\%$ confidence intervals (using 2000 test episodes).}
\label{tab:RotNetAuxMiniImagenetResults}
\vspace{-10pt}
\end{table}
\begin{table}[t]
\centering
\renewcommand{\figurename}{Table}
\renewcommand{\captionlabelfont}{\bf}
\renewcommand{\captionfont}{\small} 
{\setlength{\extrarowheight}{2pt}\footnotesize
{
\begin{tabular}{>{\hspace{-0.2em}} l <{\hspace{-0.2em}} | >{\hspace{-0.2em}} c <{\hspace{-0.2em}} | >{\hspace{-0.2em}} l <{\hspace{-0.2em}} | >{\hspace{-0.2em}} l <{\hspace{-0.2em}}}
\toprule
\multicolumn{1}{l|}{Models} & Backbone & \multicolumn{1}{c|}{1-shot} & \multicolumn{1}{c}{5-shot}\\
\midrule
\; CC       & \multirow{2}{*}{Conv-4-64} & 61.80 $\pm$ 0.30\% & 78.02 $\pm$ 0.24\%\\
\; CC+rot    &                            & \textbf{63.45} $\pm$ 0.31\% & \textbf{79.79} $\pm$ 0.24\%\\
\midrule
\; CC       &  \multirow{2}{*}{Conv-4-512}& 65.26 $\pm$ 0.31\% & 81.14 $\pm$ 0.23\%\\
\; CC+rot   &                             & \textbf{65.87} $\pm$ 0.30\% & \textbf{81.92} $\pm$ 0.23\%\\
\midrule
\; CC       & \multirow{2}{*}{WRN-28-10}  &   71.83 $\pm$ 0.31\% & 84.63 $\pm$ 0.23\%\\
\; CC+rot    &                             &  \textbf{73.62} $\pm$ 0.31\% & \textbf{86.05} $\pm$ 0.22\%\\
\midrule
\; PN       & \multirow{2}{*}{Conv-4-64}  &  62.82 $\pm$ 0.32\% & 79.59 $\pm$ 0.24\%\\
\; PN+rot   &                             &  \textbf{64.69} $\pm$ 0.32\% & \textbf{80.82} $\pm$ 0.24\%\\
\midrule
\; PN       & \multirow{2}{*}{Conv-4-512}  &  66.48 $\pm$ 0.32\% & 80.28 $\pm$ 0.23\%\\
\; PN+rot   &                              &  \textbf{67.94} $\pm$ 0.31\% & \textbf{82.20} $\pm$ 0.23\%\\
\midrule
\; PN       & \multirow{2}{*}{WRN-28-10}  &  68.35 $\pm$ 0.34\% & \textbf{81.79} $\pm$ 0.23\%\\
\; PN+rot    &                             &  \textbf{68.60} $\pm$ 0.34\% & 81.25 $\pm$ 0.24\%\\
\bottomrule
\end{tabular}}}
\caption{\textbf{Rotation prediction as auxiliary loss on CIFAR-FS.} Average 5-way classification accuracies for the novel classes on the test set of CIFAR-FS with $95\%$ confidence intervals (using 5000 test episodes).}
\label{tab:RotNetAuxCIFARFS}
\vspace{-10pt}
\end{table}
\begin{table}[t]
\centering
\renewcommand{\figurename}{Table}
\renewcommand{\captionlabelfont}{\bf}
\renewcommand{\captionfont}{\small}
{\setlength{\extrarowheight}{2pt}\footnotesize
{
\begin{tabular}{ >{\hspace{-0.2em}} l <{\hspace{-0.2em}} | >{\hspace{-0.2em}} c <{\hspace{-0.2em}} | >{\hspace{-0.2em}} l <{\hspace{-0.2em}} | >{\hspace{-0.2em}} l <{\hspace{-0.2em}} }
\toprule
\multicolumn{1}{l|}{Model} & Backbone & \multicolumn{1}{c|}{1-shot} & \multicolumn{1}{c}{5-shot}\\
\midrule
\; CC       & \multirow{2}{*}{Conv-4-64} &  53.72 $\pm$ 0.42\% & 70.96 $\pm$ 0.33\%\\
\; CC+loc   &                            &  \textbf{54.30} $\pm$ 0.42\% & \textbf{71.58} $\pm$ 0.33\%\\
\midrule
\; CC & \multirow{2}{*}{Conv-4-512} & 55.58 $\pm$ 0.42\% & 73.52 $\pm$ 0.33\%\\
\; CC+loc &                             & \textbf{56.87} $\pm$ 0.42\% & \textbf{74.84} $\pm$ 0.33\%\\
\midrule
\; CC & \multirow{2}{*}{WRN-28-10}  & 58.43 $\pm$ 0.46\% & 75.45 $\pm$ 0.34\%\\
\; CC+loc &                            & \textbf{60.71} $\pm$ 0.46\% & \textbf{77.64} $\pm$ 0.34\%\\
\bottomrule
\end{tabular}}}
\caption{\textbf{Relative patch location as auxiliary loss on MiniImageNet.} Average 5-way classification accuracies for the novel classes on the test set of MiniImageNet with $95\%$ confidence intervals (using 2000 test episodes).}
\label{tab:PatchAuxMiniImagenetResults}
\vspace{-12pt}
\end{table}

\paragraph{Rotation prediction as auxiliary loss function.}
We first study the impact of adding rotation prediction as self-supervision to the few-shot learning algorithms of Cosine Classifiers (CC) and Prototypical Networks (PN). 
We perform this study using the MiniImageNet and CIFAR-FS datasets and report results in Tables~\ref{tab:RotNetAuxMiniImagenetResults} and~\ref{tab:RotNetAuxCIFARFS} respectively.
For the CC case, we use as strong baselines, CC models without self-supervision but trained to recognize all the 4 rotated versions of an image.
The reason for using this baseline is that during the first learning stage, the model ``sees'' the same type of data, \ie, rotated images, as the model with rotation prediction self-supervision.
Note that despite the rotation augmentations of the first learning stage, during the second stage the model \emph{uses as training examples for the novel classes only the up-right versions of the images}.
Still however, using rotation augmentations improves the classification performance of the baseline models when adapted to the novel classes.
Therefore, for fair comparison, we also apply rotation augmentations to the CC models with rotation prediction self-supervision.
For the PN case, we do not use rotation augmentation since in our experiments this lead to performance degradation.

The results in Tables~\ref{tab:RotNetAuxMiniImagenetResults} and~\ref{tab:RotNetAuxCIFARFS} 
demonstrate that 
\textbf{(1)} indeed, adding rotation prediction self-supervision improves the few-shot classification performance, and 
\textbf{(2)} the performance improvement is more significant for high capacity architectures, \eg, WRN-28-10.

\paragraph{Relative patch location prediction as auxiliary loss function.}

As explained in \S\ref{sec:aux_losses}, we consider a second self-supervised task, namely relative patch location prediction.
For simplicity, we restrict its assessment to the CC few-shot algorithm, which in our experiments proved to perform better than PN and to be simpler to train.
Also, for this study we consider only the MiniImageNet dataset and not CIFAR-FS 
since the latter contains thumbnail images of size $32 \times 32$ from which it does not make sense to extract patches: their size would have to be less than $8 \times 8$ pixels, which is too small for any of the evaluated architectures. 
We report results on MiniImageNet in Table~\ref{tab:PatchAuxMiniImagenetResults}.
As a strong baseline we used CC models without self-supervision but with the auxiliary patch based classification loss described in \S\ref{sec:impl_details}.

Based on the results of Table~\ref{tab:PatchAuxMiniImagenetResults} we observe that:
\textbf{(1)}
relative patch location also manages to improve the few-shot classification performance and,
as in the rotation prediction case, the improvement is more significant for high capacity network architectures.
\textbf{(2)} Also, comparing to the rotation prediction case, the relative patch location offers smaller performance improvement. 
\hspace{-5pt}

\paragraph{Comparison with prior work.}
In Tables~\ref{tab:ComparisonWithPriorMiniImagenetResults},~\ref{tab:ComparisonWithPriorCIFARFS}, and~\ref{tab:RotationAuxTieredMiniImagenet},
we compare our approach with prior few-shot methods on the MiniImageNet, CIFAR-FS, and \emph{tiered}-MiniImageNet datasets respectively.
For our approach we used CC and rotation prediction self-supervision, which before gave the best results.
In all cases we achieve state-of-the-art results surpassing prior methods with a significant margin.
For instance, in the 1-shot and 5-shot settings of MiniImageNet we outperform the previous leading method LEO~\cite{rusu2018meta} by around $1.3$ and $2.3$ percentage points respectively.

More detailed results are provided in Appendix~\ref{sec:appendix_extra_experiments}.

\begin{table}[t]
\centering
\renewcommand{\figurename}{Table}
\renewcommand{\captionlabelfont}{\bf}
\renewcommand{\captionfont}{\small}
{\setlength{\extrarowheight}{2pt}\footnotesize
{
\begin{tabular}{ >{\hspace{-0.4em}} l <{\hspace{-0.4em}} | >{\hspace{-0.4em}} l <{\hspace{-0.4em}}| >{\hspace{-0.4em}} l <{\hspace{-0.4em}} | >{\hspace{-0.4em}} l <{\hspace{-0.4em}}}
\toprule
\multicolumn{1}{l|}{Models} & Backbone & \multicolumn{1}{c|}{1-shot} & \multicolumn{1}{c}{5-shot}\\
\midrule
\;MAML~\cite{finn2017model}                      & Conv-4-64  & 48.70 $\pm$ 1.84\% & 63.10 $\pm$ 0.92\%\\
\;Prototypical Nets~\cite{snell2017prototypical} & Conv-4-64  & 49.42 $\pm$ 0.78\% & 68.20 $\pm$ 0.66\%\\
\;LwoF~\cite{gidaris2018dynamic}                 & Conv-4-64  & 56.20 $\pm$ 0.86\% & 72.81 $\pm$ 0.62\%\\
\;RelationNet~\cite{yang2018learning}            & Conv-4-64  & 50.40 $\pm$ 0.80\% & 65.30 $\pm$ 0.70\%\\
\;GNN~\cite{garcia2017few}                       & Conv-4-64  & 50.30\%          & 66.40\%\\
\;R2-D2~\cite{bertinetto2018meta}                & Conv-4-64  & 48.70 $\pm$ 0.60\% & 65.50 $\pm$ 0.60\%\\
\;R2-D2~\cite{bertinetto2018meta}                & Conv-4-512 & 51.20 $\pm$ 0.60\% & 68.20 $\pm$ 0.60\%\\
\;TADAM~\cite{oreshkin2018tadam}                 & ResNet-12      & 58.50 $\pm$ 0.30\% & 76.70 $\pm$ 0.30\%\\
\;Munkhdalai~\etal\cite{munkhdalai2017meta}      & ResNet-12      & 57.10 $\pm$ 0.70\% & 70.04 $\pm$ 0.63\%\\
\;SNAIL~\cite{santoro2017simple}                 & ResNet-12      & 55.71 $\pm$ 0.99\% & 68.88 $\pm$ 0.92\%\\
\;Qiao~\etal\cite{qiao2017few}$^\ast$                & WRN-28-10 & 59.60 $\pm$ 0.41\% & 73.74 $\pm$ 0.19\%\\
\;LEO~\cite{rusu2018meta}$^\ast$                     & WRN-28-10 & 61.76 $\pm$ 0.08\% & 77.59 $\pm$ 0.12\%\\
\midrule
\; CC+rot                              & Conv-4-64  & 54.83 $\pm$ 0.43\% & 71.86 $\pm$ 0.33\%\\
\; CC+rot                              & Conv-4-512 & 56.27 $\pm$ 0.43\% & 74.30 $\pm$ 0.34\%\\
\; CC+rot                              & WRN-28-10  & \textbf{62.93} $\pm$ 0.45\% & \textbf{79.87} $\pm$ 0.33\%\\
\midrule
\; CC+rot+unlabeled                    & WRN-28-10  & \textbf{63.77} $\pm$ 0.45\% & \textbf{80.70} $\pm$ 0.33\%\\
\bottomrule
\end{tabular}}}
\caption{\textbf{Comparison with prior work on MiniImageNet.} Average 5-way classification accuracies for the novel classes on the test set of MiniImageNet with $95\%$ confidence intervals (using 2000 test episodes).
$^\ast$: using also the validation classes for training.
For the description of the CC+rot+unlabeled model see~\S\ref{sec:semi_experiments}.
}
\vspace{-8pt}
\label{tab:ComparisonWithPriorMiniImagenetResults}
\end{table}
\begin{table}[t]
\centering
\renewcommand{\figurename}{Table}
\renewcommand{\captionlabelfont}{\bf}
\renewcommand{\captionfont}{\small}
{\setlength{\extrarowheight}{2pt}\footnotesize
{
\begin{tabular}{>{\hspace{-0.4em}} l <{\hspace{-0.4em}} | >{\hspace{-0.4em}} l <{\hspace{-0.4em}} | >{\hspace{-0.4em}} l | >{\hspace{-0.4em}} l <{\hspace{-0.4em}}}
\toprule
\multicolumn{1}{l|}{Models} & Backbone & \multicolumn{1}{c|}{1-shot} & \multicolumn{1}{c}{5-shot}\\
\midrule
\;PN~\cite{snell2017prototypical}$^\dagger$ & Conv-4-64  & 55.50 $\pm$ 0.70\% & 72.00 $\pm$ 0.60\%\\
\;PN~\cite{snell2017prototypical}$^\dagger$ & Conv-4-512 & 57.90 $\pm$ 0.80\% & 76.70 $\pm$ 0.60\%\\
\;PN~\cite{snell2017prototypical}$^\ddagger$                     & Conv-4-64  & 62.82 $\pm$ 0.32\% & 79.59 $\pm$ 0.24\%\\
\;PN~\cite{snell2017prototypical}$^\ddagger$                     & Conv-4-512 & 66.48 $\pm$ 0.32\% & 80.28 $\pm$ 0.23\%\\
\;MAML~\cite{finn2017model}$^\dagger$                       & Conv-4-64  & 58.90 $\pm$ 1.90\% & 71.50 $\pm$ 1.00\%\\
\;MAML~\cite{finn2017model}$^\dagger$                       & Conv-4-512 & 53.80 $\pm$ 1.80\% & 67.60 $\pm$ 1.00\%\\
\;RelationNet~\cite{yang2018learning}$^\dagger$             & Conv-4-64  & 55.00 $\pm$ 1.00\% & 69.30 $\pm$ 0.80\%\\
\;GNN~\cite{garcia2017few}$^\dagger$                        & Conv-4-64  & 61.90\%            & 75.30\%\\
\;GNN~\cite{garcia2017few}$^\dagger$                        & Conv-4-512 & 56.00\%            & 72.50\%\\
\;R2-D2~\cite{bertinetto2018meta}                               & Conv-4-64  & 60.00 $\pm$ 0.70\% & 76.10 $\pm$ 0.60\%\\
\;R2-D2~\cite{bertinetto2018meta}                               & Conv-4-512 & 64.00 $\pm$ 0.80\% & 78.90 $\pm$ 0.60\%\\
\midrule
\; CC+rot                        & Conv-4-64  & 63.45 $\pm$ 0.31\% & 79.79 $\pm$ 0.24\%\\
\; CC+rot                        & Conv-4-512 & 65.87 $\pm$ 0.30\% & 81.92 $\pm$ 0.23\%\\
\; CC+rot                        & WRN-28-10  & \textbf{73.62} $\pm$ 0.31\% & \textbf{86.05} $\pm$ 0.22\%\\
\bottomrule
\end{tabular}}}
\caption{\textbf{Comparison with prior work in CIFAR-FS.} Average 5-way classification accuracies for the novel classes on the test set of CIFAR-FS with $95\%$ confidence (using 5000 test episodes).
$^\dagger$: results from~\cite{bertinetto2018meta}.
$^\ddagger$: our implementation.
}
\label{tab:ComparisonWithPriorCIFARFS}
\vspace{-8pt}
\end{table}
\begin{table}[t]
\centering
\renewcommand{\figurename}{Table}
\renewcommand{\captionlabelfont}{\bf}
\renewcommand{\captionfont}{\small}
{\setlength{\extrarowheight}{2pt}\footnotesize
{
\begin{tabular}{>{\hspace{-0.4em}} l <{\hspace{-0.4em}} | >{\hspace{-0.4em}} l <{\hspace{-0.4em}} | >{\hspace{-0.4em}} l <{\hspace{-0.4em}} | >{\hspace{-0.4em}} l <{\hspace{-0.4em}}}
\toprule
\multicolumn{1}{l|}{Models} & Backbone & \multicolumn{1}{c|}{1-shot} & \multicolumn{1}{c}{5-shot}\\
\midrule
\;MAML~\cite{finn2017model}$^\dagger$  & Conv-4-64          &  51.67 $\pm$ 1.81\% &  70.30 $\pm$ 0.08\% \\
\;Prototypical Nets~\cite{snell2017prototypical}  & Conv-4-64                       &  53.31 $\pm$ 0.89\% &  72.69 $\pm$ 0.74 \% \\
\;RelationNet~\cite{yang2018learning}$^\dagger$ & Conv-4-64 &  54.48 $\pm$ 0.93\% &  71.32 $\pm$ 0.78\%\\
\;Liu~\etal~\cite{liu2018transductive} & Conv-4-64                                 &  57.41 $\pm$ 0.94\% & 71.55 $\pm$ 0.74\\
\;LEO~\cite{rusu2018meta}   & WRN-28-10                                             &  66.33 $\pm$ 0.05\% &  81.44 $\pm$ 0.09 \%\\
\midrule
\; CC                        & WRN-28-10 & 70.04 $\pm$ 0.51\% & 84.14 $\pm$ 0.37\% \\
\; CC+rot                    & WRN-28-10 & \textbf{70.53} $\pm$ 0.51\% & \textbf{84.98} $\pm$ 0.36\% \\
\bottomrule
\end{tabular}}}
\caption{\textbf{Rotation prediction as auxiliary loss on \emph{tiered}-MiniImageNet.} Average 5-way classification accuracies for the novel classes on the test set of \emph{tiered}-MiniImageNet with $95\%$ confidence (using 2000 test episodes for our entries). 
$^\dagger$: results from~\cite{liu2018transductive}.}
\vspace{-8pt}
\label{tab:RotationAuxTieredMiniImagenet}
\end{table}

\subsection{Semi-supervised few-shot learning} \label{sec:semi_experiments}

Next, we evaluate the proposed semi-supervised training approach.
In these experiments we use CC models with rotation prediction self-supervision.
We perform two types of semi-supervised experiments: 
(1) training with unlabeled data from the same base classes, and 
(2) training with unlabeled data that are not from the base classes.

\paragraph{Training with unlabeled data from the same base classes.}
From the base classes of MiniImageNet, we use only a small percentage of the training images (\eg, 5\% of images per class) as annotated training data, while the rest of the images (\eg, 95\%) are used as the unlabeled data in the semi-supervised training.
We provide results in the first two sections of Table~\ref{tab:SemiRotationMiniImagenetResults}.
The proposed semi-supervised training approach is compared with a CC model without self-supervision and with a CC model with self-supervision but no recourse to the unlabeled data. 
The results demonstrate that indeed, our method manages to improve few-shot classification performance by exploiting unlabeled images.
Compared to Ren~\etal~\cite{ren2018meta}, that also propose a semi-supervised method,
our method with Conv-4-64 and 20\% annotations achieves better results than their method with Conv-4-64 and 40\% annotations 
(\ie, our \textbf{51.21}\% and \textbf{68.89}\% MiniImageNet accuracies vs. their 50.41\% and 64.39\% for the 1-shot and 5-shot settings respectively).

\paragraph{Training with unlabeled data not from the base classes.}
This is a more realistic setting, since it is hard to constrain the unlabeled images to be from the same classes as the base classes.
For this experiment, we used as unlabeled data the training images of the \emph{tiered}-MiniImageNet base classes \emph{minus the classes that are common with the base, validation, or test classes of MiniImageNet}.
In total, $408,726$ unlabeled images are used from \emph{tiered}-MiniImageNet.
We report results in the last row of Table~\ref{tab:SemiRotationMiniImagenetResults}.
Indeed, even in this difficult setting, our semi-supervised approach is still able to exploit unlabeled data and improve the classification performance. 
Furthermore, we did an extra experiment in which we trained a WRN-28-10 based model using 100\% of MiniImageNet training images and unlabeled data from \emph{tiered}-MiniImageNet. 
This model achieved \textbf{63.77\%} and \textbf{80.70\%} accuracies for the 1-shot and 5-shot settings respectively on MiniImageNet (see entry CC+rot+unlabeled of Table~\ref{tab:ComparisonWithPriorMiniImagenetResults}), 
which improves over the already very strong CC+rot model (see Table~\ref{tab:ComparisonWithPriorMiniImagenetResults}).

\subsection{Few-shot object recognition to assess self-supervised representations} \label{sec:self_experiments}
\begin{table}[t]
\centering
\renewcommand{\figurename}{Table}
\renewcommand{\captionlabelfont}{\bf}
\renewcommand{\captionfont}{\small}
{\setlength{\extrarowheight}{2pt}\footnotesize
{
\begin{tabular}{ >{\hspace{-0.4em}} c <{\hspace{-0.4em}} | >{\hspace{-0.4em}} c <{\hspace{-0.4em}} | >{\hspace{-0.4em}} c <{\hspace{-0.4em}} | >{\hspace{-0.5em}} l <{\hspace{-0.5em}} | >{\hspace{-0.5em}} l <{\hspace{-0.5em}} | >{\hspace{-0.5em}} l <{\hspace{-0.5em}} | >{\hspace{-0.5em}} l <{\hspace{-0.5em}} | >{\hspace{-0.5em}} l <{\hspace{-0.5em}} | >{\hspace{-0.5em}} l <{\hspace{-0.5em}}}
\toprule
Rot & M & T & \multicolumn{2}{c|}{$\mu=5$\%} & \multicolumn{2}{c|}{$\mu=10$\%} & \multicolumn{2}{c}{$\mu=20$\%}\\
& & & \multicolumn{1}{c|}{1-shot} & \multicolumn{1}{c|}{5-shot} & \multicolumn{1}{c|}{1-shot} & \multicolumn{1}{c|}{5-shot} & \multicolumn{1}{c|}{1-shot} & \multicolumn{1}{c}{5-shot}\\
\midrule
\multicolumn{3}{>{\hspace{-0.4em}} c <{\hspace{-0.4em}}|}{Conv-4-64} & \multicolumn{6}{c}{}\\
           &            &            & 41.87\% & 57.76\% & 45.63\% & 62.29\% & 49.34\% & 66.48\%\\
\checkmark &            &            & 43.26\% & 58.88\% & 47.57\% & 64.03\% & 50.48\% & 67.92\%\\
\checkmark & \checkmark &            & \textbf{45.41}\% & \textbf{62.14}\% & \textbf{47.75}\% & \textbf{64.93}\% & \textbf{51.21}\% & \textbf{68.89}\%\\
\midrule
\multicolumn{3}{>{\hspace{-0.4em}} c <{\hspace{-0.4em}} |}{WRN-28-10} & \multicolumn{6}{c}{}\\
           &            &            & 39.05\% & 54.89\% & 44.59\% & 61.42\% & 49.52\% & 67.81\%\\
\checkmark &            &            & 43.60\% & 60.52\% & 49.05\% & 67.35\% & 53.66\% & 72.69\%\\
\checkmark & \checkmark &            & \textbf{47.25}\% & \textbf{65.07}\% & \textbf{52.90}\% & \textbf{71.48}\% & \textbf{56.79}\% & \textbf{74.67}\%\\
\midrule
\checkmark &            & \checkmark & 46.95\% & 64.93\% & 52.66\% & 71.13\% & 55.37\% & 73.66\%\\
\bottomrule
\end{tabular}}}
\caption{\textbf{Semi-supervised training with rotation prediction as self-supervision on MiniImageNet.} Average 5-way classification accuracies on the test set of MiniImageNet. 
$\mu$ is the percentage of the base class training images of MiniImageNet that are used as annotated data during training.
\emph{Rot} indicates adding self-supervision,
\emph{M} indicates using as unlabeled data the (rest of) MiniImageNet training dataset, and
\emph{T} using as unlabeled data the \emph{tiered}-MiniImageNet training dataset.}
\vspace{-8pt}
\label{tab:SemiRotationMiniImagenetResults}
\end{table}
\begin{table}[t]
\centering
\renewcommand{\figurename}{Table}
\renewcommand{\captionlabelfont}{\bf}
\renewcommand{\captionfont}{\small} 
{\setlength{\extrarowheight}{2pt}\footnotesize
{
\begin{tabular}{>{\hspace{-0.5em}} l <{\hspace{-0.5em}} | >{\hspace{-0.5em}} c <{\hspace{-0.5em}}| >{\hspace{-0.5em}} c <{\hspace{-0.5em}} | >{\hspace{-0.5em}} c <{\hspace{-0.5em}} | >{\hspace{-0.5em}} c <{\hspace{-0.5em}} | >{\hspace{-0.5em}} c <{\hspace{-0.5em}}}
\toprule
\multicolumn{1}{l|}{Models} & Backbone & \multicolumn{1}{c|}{1-shot} & \multicolumn{1}{c|}{5-shot} & \multicolumn{1}{c|}{20-shot} & \multicolumn{1}{c}{50-shot}\\
\midrule
\;CACTUs~\cite{hsu2019unsupervised}   & Conv-4-64 & 39.90\% & 53.97\% & 63.84\% & 69.64\%\\
\midrule
\;CC                                  &  \multirow{3}{*}{Conv-4-64} & 53.63\% & 70.74\% & 80.03\% & 82.61\%\\
\;Rot                                 &                             & 41.70\% & 58.64\% & 68.61\% & 71.86\%\\
\;Loc                                 &                             & 37.75\% & 53.02\% & 61.38\% & 64.15\%\\
\midrule
\;CC                                 & \multirow{3}{*}{WRN-28-10} & 58.59\% & 76.59\% & 82.70\% & 84.27\%\\
\;Rot                                &                            & 43.43\% & 60.86\% & 69.82\% & 72.20\%\\
\;Loc                                &                            & 41.78\% & 59.10\% & 67.86\% & 70.32\%\\
\bottomrule
\end{tabular}}}
\caption{
\textbf{Evaluating self-supervised representation learning methods on few-shot recognition.} 
Average 5-way classification accuracies on the test set of MiniImageNet.
\emph{Rot} refers to the rotation prediction task, \emph{Loc} to the relative patch location task, and \emph{CC} to the supervised method of Cosine Classifiers.}
\vspace{-8pt}
\label{tab:UnsupervisedMiniImagenetResults}
\end{table}

Given that our framework allows the easy combination of any type of self-supervised learning approach with the adopted few-shot learning algorithms, we also propose to use it as an alternative way for comparing/evaluating the effectiveness of different self-supervised approaches.
To this end, the only required change to our framework is to use uniquely the self-learning loss (\ie, no labeled data is now used) in the first learning stage (for implementation details see Appendix~\ref{sec:appendix_selfsupervision_training}). The performance of the few-shot model resulting from the second learning stage can then be used for evaluating the self-supervised method under consideration.

Comparing competing self-supervised techniques is not straightforward since it must be done by setting up another, somewhat contrived task that exploits the learned representations \cite{doersch2015unsupervised, kolesnikov2019revisiting}.
Instead, given the very similar goals of few-shot and self-supervised learning,
we argue that the proposed comparison method could be more meaningful for assessing different self-supervised features.
Furthermore, it is quite simple and fast to perform when compared to some alternatives such as fine-tuning the learned representations on the PASCAL~\cite{Everingham10} detection task~\cite{doersch2015unsupervised}, 
with the benefit of obtaining more robust statistics 
aggregated over evaluations of thousands of episodes with multiple different configurations of classes and training/testing samples.

To illustrate our point, we provide in Table~\ref{tab:UnsupervisedMiniImagenetResults} quantitative results of this type of evaluation on the MiniImageNet dataset, for the self-supervision methods of rotation prediction and relative patch location prediction. For self-supervised training we used the training images of the base classes of MiniImageNet and for the few-shot classification step we used the test classes of MiniImageNet. 
We observe that the explored self-supervised approaches achieve relatively competitive classification performance when compared to the supervised method of CC and obtain results that are on par or better than other, more complex, unsupervised systems.
We leave as future work a more detailed and thorough comparison of self-learned representations in this evaluation setting.

%% file: conclusions.tex
\section{Conclusions} \label{sec:conclusions}

Inspired by the close connection between few-shot and self-supervised learning, we propose to add an auxiliary loss based on self-supervision during the training of few-shot recognition models.
The goal is to boost the ability of the latter to recognize novel classes using only few training data. 
Our detailed experiments on \mini, \fscifar, and \tmini~few-shot datasets reveal that indeed adding self-supervision leads to significant improvements on the few-shot classification performance, which makes the employed few-shot models achieve state-of-the-art results. 
Furthermore, the annotation-free nature of the self-supervised loss allows us to exploit diverse unlabeled data in a semi-supervised manner, which further improves the classification performance. 
Finally, we show that the proposed framework can also be used for evaluating self-supervised or unsupervised methods based on few-shot object recognition.

%% file: appendix.tex
\section{Extra experimental results} \label{sec:appendix_extra_experiments}

\subsection{Rotation prediction self-supervision: Impact of rotation augmentation} \label{sec:appendix_rotation_experiments}

In the experiments reported in Section \ref{sec:experiments}, we use rotation augmentation when training the baselines to compare against the CC-models with self-supervised rotation prediction. 
In Table~\ref{tab:AppendixRotNetAuxMiniImagenetResults} of this Appendix we also provide results without using rotation augmentation.
The purpose is to examine what is the impact of this augmentation technique.
We observe that (1) the improvements yielded by rotation prediction self-supervision are more significant, and (2) in some cases rotation augmentation actually reduces the few-shot classification performance.

\subsection{Relative patch location self-supervision: impact of patch based object classification loss} \label{sec:appendix_location_experiments}

When we study the impact of adding relative patch location self-supervision to CC-based models in Section \ref{sec:experiments}, we use an auxiliary patch based object classification loss. 
In Table~\ref{tab:AppendixPatchAuxMiniImagenetResults} we also provide results without using this auxiliary loss when training CC models.
The purpose is to examine what is the impact of this auxiliary loss.
We observe that the improvement brought by this auxiliary loss is small (or no-existing) when compared to the performance improvement thanks to the relative patch location self-supervision.

\begin{table*}[t]
\centering
\renewcommand{\figurename}{Table}
\renewcommand{\captionlabelfont}{\bf}
\renewcommand{\captionfont}{\small}
{\setlength{\extrarowheight}{2pt}\footnotesize
{
\begin{tabular}{l  |  c |  c |  l |  l |  l | l}
\toprule
\multirow{2}{*}{Model} & \multirow{2}{*}{Rot. Aug.} & \multirow{2}{*}{Backbone} & \multicolumn{2}{c|}{MiniImageNet} & \multicolumn{2}{c}{CIFAR-FS}\\
&  &  & \multicolumn{1}{c|}{1-shot} & \multicolumn{1}{c|}{5-shot} & \multicolumn{1}{c|}{1-shot} & \multicolumn{1}{c}{5-shot}\\
\midrule
\; CC       &            & \multirow{4}{*}{Conv-4-64} & 53.94 $\pm$ 0.42\% & 71.13 $\pm$ 0.34\% & 62.83 $\pm$ 0.31\% & 79.14 $\pm$ 0.24\%\\
\; CC+rot   &            &                            & \textbf{55.41} $\pm$ 0.43\% & \textbf{72.98} $\pm$ 0.33\% & \textbf{63.98} $\pm$ 0.31\% & \textbf{80.44} $\pm$ 0.23\%\\
\; CC       & \checkmark &                            & 54.31 $\pm$ 0.42\% & 70.89 $\pm$ 0.34\% & 61.80 $\pm$ 0.30\% & 78.02 $\pm$ 0.24\%\\
\; CC+rot   & \checkmark &                            & 54.83 $\pm$ 0.43\% & 71.86 $\pm$ 0.33\% & 63.45 $\pm$ 0.31\% & 79.79 $\pm$ 0.24\%\\
\midrule
\; CC       &            & \multirow{4}{*}{Conv-4-512}  & 54.51 $\pm$ 0.42\% & 72.52 $\pm$ 0.34\% & 65.64 $\pm$ 0.31\% & 81.10 $\pm$ 0.23\%\\
\; CC+rot   &            &                              & \textbf{56.59} $\pm$ 0.43\% & \textbf{74.67} $\pm$ 0.34\% & \textbf{67.00} $\pm$ 0.30\% & \textbf{82.55} $\pm$ 0.23\%\\
\; CC       & \checkmark &                              & 55.68 $\pm$ 0.43\% & 73.19 $\pm$ 0.33\% & 65.26 $\pm$ 0.31\% & 81.14 $\pm$ 0.23\%\\
\; CC+rot   & \checkmark &                              & 56.27 $\pm$ 0.43\% & 74.30 $\pm$ 0.33\% & 65.87 $\pm$ 0.30\% & 81.92 $\pm$ 0.23\%\\
\midrule
\; CC       &            & \multirow{4}{*}{WRN-28-10} &  58.59 $\pm$ 0.45\% & 76.59 $\pm$ 0.33\% & 70.43 $\pm$ 0.31\% & 83.84 $\pm$ 0.23\%\\
\; CC+rot   &            &                            &  60.10 $\pm$ 0.45\% & 77.40 $\pm$ 0.33\% & 72.49 $\pm$ 0.31\% & 84.77 $\pm$ 0.22\%\\
\; CC       & \checkmark &                            &  61.09 $\pm$ 0.44\% & 78.43 $\pm$ 0.33\% & 71.83 $\pm$ 0.31\% & 84.63 $\pm$ 0.23\%\\\
\;CC+rot   & \checkmark &                            &  \textbf{62.93} $\pm$ 0.45\% & \textbf{79.87} $\pm$ 0.33\% & \textbf{73.62} $\pm$ 0.31\% & \textbf{86.05} $\pm$ 0.22\%\\
\bottomrule
\end{tabular}}}
\caption{\textbf{Impact of rotation augmentation.} Average 5-way classification accuracies for the novel classes on the test sets of MiniImageNet and CIFAR-FS with $95\%$ confidence intervals.
\emph{Rot. Aug.} indicates using rotation augmentation during the first learning stage.
}
\label{tab:AppendixRotNetAuxMiniImagenetResults}
\vspace{-8pt}
\end{table*}

\begin{table*}[t]
\centering
\renewcommand{\figurename}{Table}
\renewcommand{\captionlabelfont}{\bf}
\renewcommand{\captionfont}{\small}
{\setlength{\extrarowheight}{2pt}\footnotesize
{
\begin{tabular}{l | c |  c |  l |  l }
\toprule
Model & Patch Cls. & Backbone  & \multicolumn{1}{c|}{1-shot} & \multicolumn{1}{c}{5-shot}\\
\midrule
\; CC      &            & \multirow{3}{*}{Conv-4-64} & 53.63 $\pm$ 0.42\% & 70.74 $\pm$ 0.34\%\\
\; CC      & \checkmark &                            & 53.72 $\pm$ 0.42\% & 70.96 $\pm$ 0.33\%\\
\; CC+loc  & \checkmark &                            & \textbf{54.30} $\pm$ 0.42\% & \textbf{71.58} $\pm$ 0.33\%\\
\midrule
\; CC     &            & \multirow{3}{*}{Conv-4-512} & 54.51 $\pm$ 0.42\% & 72.52 $\pm$ 0.34\%\\
\; CC     & \checkmark &                             & 55.58 $\pm$ 0.42\% & 73.52 $\pm$ 0.33\%\\
\; CC+loc & \checkmark &                             & \textbf{56.87} $\pm$ 0.42\% & \textbf{74.84} $\pm$ 0.33\%\\
\midrule
\; CC     &            & \multirow{3}{*}{WRN-28-10} & 58.59 $\pm$ 0.45\% & 76.59 $\pm$ 0.33\%\\
\; CC     & \checkmark &                            & 58.43 $\pm$ 0.46\% & 75.45 $\pm$ 0.34\%\\
\; CC+loc & \checkmark &                            & \textbf{60.71} $\pm$ 0.46\% & \textbf{77.64} $\pm$ 0.34\%\\
\bottomrule
\end{tabular}}}
\caption{\textbf{Impact of auxiliary patch based object classification loss.} Average 5-way classification accuracies for the novel classes on the test set of MiniImageNet with $95\%$ confidence intervals.
\emph{Patch Cls.} indicates using an auxiliary patch based object classification loss during the first learning stage.
}
\label{tab:AppendixPatchAuxMiniImagenetResults}
\vspace{-8pt}
\end{table*}

\section{Additional implementation details} \label{sec:appendix_implementation_details}

\subsection{Network architectures} \label{sec:appendix_network_architectures}

\paragraph{Conv-4-64~\cite{vinyals2016matching}.}
It consists of $4$ convolutional blocks each implemented with a $3 \times 3$ convolutional layer with $64$ channels followed by BatchNorm + ReLU + $2 \times 2$ max-pooling units.
In the MiniImageNet experiments for which the image size is $84 \times 84$ pixels, its output feature map has size $5 \times 5 \times 64$ and is flattened into a final $1600$-dimensional feature vector. For the CIFAR-FS experiments, the image size is $32 \times 32$ pixels, the output feature map has size $2 \times 2 \times 64$ and is flattened into a $256$-dimensional feature vector.

\paragraph{Conv-4-512.} It is derived from Conv-4-64 by gradually increasing its width across layers leading to $96$, $128$, $256$, and $512$ feature channels for its 4 convolutional blocks respectively. 
Therefore, for a $84 \times 84$ sized image (\ie, MiniImageNet experiments) its output feature map has size $5 \times 5 \times 512$ and is flattened into a final $12800$-dimensional feature vector, while for a $32 \times 32$ sized image (\ie, CIFAR-FS experiments) its output feature map has size $2 \times 2 \times 512$ and is flattened into a final $2048$-dimensional feature vector.

\paragraph{WRN-28-10~\cite{zagoruyko2016wide}.}
It is a Wide Residual Network with $28$ convolutional layers and width factor $10$.
The $12$ residual layers of this architecture are grouped into $3$ residual blocks ($4$ residual layers per block).
In the MiniImageNet and \emph{tiered}-MiniImageNet experiments, 
the network gets as input images of size $80 \times 80$ (rescaled from $84 \times 84$), 
and during feature extraction each residual block downsamples by a factor of 2 the processed feature maps.
Therefore, the output feature map has size $10 \times 10 \times 640$ which, after global average pooling, creates a $640$-dimensional feature vector.
In the CIFAR-FS experiments, the input images have size $32 \times 32$ and during feature extraction only the last two residual blocks downsample the processed feature maps.
Therefore, in the CIFAR-FS experiments, the output feature map has size $8 \times 8 \times 640$ which again after global average pooling creates a $640$-dimensional feature vector. 

\paragraph{Rotation prediction network, $R_{\phi}(\cdot)$.}
This network gets as input the output feature maps of $F_{\theta}$ and is implemented as a convnet.
More specifically, for the Conv-4-64 and Conv-4-512 feature extractor architectures (regardless of the dataset), 
$R_{\phi}$ consists of two $3 \times 3$ convolutional layers with BatchNorm + ReLU units, followed by a fully connected classification layer. 
For Conv-4-64, those two convolutional layers have $128$ and $256$ feature channels respectively, while for Conv-4-512 both convolutional layers have $512$ feature channels. 
In the WRN-28-10 case, $R_{\phi}$ consists of a $4$-residual-layer residual block that actually replicates the last (3rd) residual block of WRN-28-10. This residual block is followed by global average pooling plus a fully connected classification layer.

\paragraph{Relative patch location network, $P_{\phi}(\cdot, \cdot)$.}
Given two patches, $P_{\phi}(\cdot, \cdot)$ gets the concatenation of their feature vectors extracted with $F_{\theta}$ as input, and forwards it to two fully connected layers.
The single hidden layer, which includes BatchNorm + ReLU units, has $256$, $1024$, and $1280$ channels for the Conv-4-64, Conv-4-512, and WRN-28-10 architectures respectively.

\subsection{Incorporating self-supervision during training} \label{sec:appendix_selfsupervision_training}

Here we provide more implementation details regarding how we incorporate self-supervision during the fist learning stage.

\paragraph{Training with rotation prediction self-supervision.}
During training for each image of a mini-batch we create its $4$ rotated copies and apply to them the rotation prediction task (\ie, $L_{\mathrm{self}}$ loss).
When training the object classifier with rotation augmentation (\eg, CC-based models) the object classification task (\ie, $L_{\mathrm{few}}$ loss) is applied to all rotated versions of the images. 
Otherwise, only the upright images (\ie, the $0$ degrees images) are used for the object classification task.
Note that in the PN-based models, 
we apply the rotation task to both the support and the query images of a training episode, and also we do not use rotation augmentation for the object classification task.

\paragraph{Training with relative patch location self-supervision.}
In this case during training each mini-batch includes two types of visual data, images and patches.
Similar to~\cite{kolesnikov2019revisiting}, 
in order to create patches, an image is:
(1) resized to $96 \times 96$ pixels (from $84 \times 84$),
(2) converted to grayscale with probability $0.66$, and then
(3) divided into $9$ regions of size $32 \times 32$ with a $3 \times 3$ regular grid.
From each $32 \times 32$ sized region we (4) randomly sample a $24 \times 24$ patch, and then (5) normalize the pixels of the patch individually to have zero mean and unit standard deviation.
The object classification task is applied to the image data of the mini-batch while the relative patch location task to the patch data of the mini-batch.
Also, as already explained, 
we also apply an extra auxiliary object classification loss to the patch data.

\subsection{Training routine for first learning stage} \label{sec:appendix_1step_training}

To optimize the training loss we use mini-batch SGD optimizer with momentum $0.9$ and weight decay $5\mathrm{e}-4$.
In the MiniImageNet and CIFAR-FS experiments, 
we train the models for $60$ epochs (each with $1000$ SGD iterations), 
starting with a learning rate of $0.1$ which is decreased by a factor of $10$ every $20$ epochs.
In the \emph{tiered}-MiniImageNet experiments we train for $100$ epochs (each with $2000$ SGD iterations), 
starting with a learning rate of $0.1$ which is decreased by a factor of $10$ every $40$ epochs.
The mini-batch sizes were cross-validated on the validation split.
For instance, the models based on CC and Conv-4-64, Conv-4-512, or WRN-28-10 architectures are trained with mini-batch sizes equal to $128$, $128$, or $64$ respectively.
Finally, we perform early stopping \wrt the few-shot classification accuracy on the validation novel classes (for the CC-based models we use the 1-shot classification accuracy).

\paragraph{Semi-supervised training.}
Here each mini-batch consists of both labeled and unlabeled data.
Specifically, for the experiments that use the Conv-4-64 network architecture, and $5\%$, $10\%$, or $20\%$ of MiniImageNet as labeled data, each mini-batch consists of $64$ labeled images and $64$ unlabeled images.
For the experiments that use the WRN-28-10 network architecture, and $5\%$, $10\%$, or $20\%$ of MiniImageNet as labeled data, each mini-batch consists of $16$ labeled images and $48$ unlabeled images.
For the experiment that uses $100\%$ of MiniImageNet as labeled data and \emph{tiered-}MiniImageNet for unlabeled data, then each mini-batch consists of $32$ labeled images and $32$ unlabeled images.

\subsection{Assessing self-supervised representations based on the few-shot object recognition task} \label{sec:appendix_evaluating_selfsupervised_representations}

Here we provide implementation details for the experiments in~\S\ref{sec:self_experiments}, 
which assess the self-supervised representations using the few-shot object recognition task.
Except from the fact that during the first learning stage,
(1) there is no object based supervision (\ie, no $L_{\mathrm{few}}$ loss), and 
(2) \emph{no early stopping based on a validation set},
the rest of the implementation details remain the same as in the other CC-based experiments.
A minor difference is that,
when evaluating the WRN-28-10 and relative patch location based model, 
we create the representation of each image by averaging the extracted feature vectors of its 9 patches (similar to~\cite{kolesnikov2019revisiting}).